\documentclass[conference]{IEEEtran}
\IEEEoverridecommandlockouts
\usepackage{cite}
\usepackage{amsmath,amssymb,amsfonts}
\usepackage{algorithmic}
\usepackage{graphicx}
\usepackage{textcomp}
\usepackage{xcolor}
\usepackage{multirow}
\def\BibTeX{{\rm B\kern-.05em{\sc i\kern-.025em b}\kern-.08em
    T\kern-.1667em\lower.7ex\hbox{E}\kern-.125emX}}
\begin{document}

\title{Low-Rank Temporal Attention-Augmented Bilinear Network for financial time-series forecasting\\

}

\author{\IEEEauthorblockN{Mostafa Shabani}
\IEEEauthorblockA{\textit{Department of Engineering} \\
\textit{Aarhus University}\\
Aarhus, Denmark \\
mshabani@eng.au.dk}
\and
\IEEEauthorblockN{Alexandros Iosifidis}
\IEEEauthorblockA{\textit{Department of Engineering} \\
\textit{Aarhus University}\\
Aarhus, Denmark \\
ai@eng.au.dk}
}

\IEEEoverridecommandlockouts
\IEEEpubid{\makebox[\columnwidth]
{978-1-7281-2547-3/20/\$31.00~\copyright2020 IEEE \hfill}
\hspace{\columnsep}\makebox[\columnwidth]{ }}
% Add the following code after the \maketitle command:
\IEEEpubidadjcol

\maketitle

\begin{abstract}
Financial market analysis, especially the prediction of movements of stock prices, is a challenging problem. The nature of financial time-series data, being non-stationary and nonlinear, is the main cause of these challenges. Deep learning models have led to significant performance improvements in many problems coming from different domains, including prediction problems of financial time-series data. Although the prediction performance is the main goal of such models, dealing with ultra high-frequency data sets restrictions in terms of the number of model parameters and its inference speed. The Temporal Attention-Augmented Bilinear network was recently proposed as an efficient and high-performing model for Limit Order Book time-series forecasting. In this paper, we propose a low-rank tensor approximation of the model to further reduce the number of trainable parameters and increase its speed.
\end{abstract}


\section{Introduction}
\label{sec:intro}
In recent years, deep learning models are achieving significant improvement in prediction problems coming from various application domains. Time series forecasting is one of the problems which is highly affected by new and powerful deep learning models. Several deep learning models have been recently applied in financial time-series analysis problems and shown performance improvements compared to traditional statistical approaches, such as the Auto-Regressive Integrated Moving Average model \cite{Box1968SomeRA, tiao1981modeling}, and methodologies based on shallow machine learning models like the K-Nearest Neighbor \cite{Dudani1976Knn} and the Support Vector Machine \cite{Huang2005} classifiers. While standard deep learning models, like Convolutional Neural Networks \cite{Zhang2019,Tsantekidis2017}, Recurrent Neural Networks \cite{Dixon} and Long-Short Term Memory Networks \cite{Tsantekidis2017a} have been proposed for financial time series analysis and outperformed prior machine learning solutions, the improvements in the obtained performance comes with an increased computational complexity which may render the use of such high-performing methodologies in ultra-high frequency financial time series forecasting impractical.

Efficient deep learning models have been recently proposed to address this issue. A neural network formulation of the Bag-of-Features model was used in \cite{Passalis2017TimeseriesCU} and was shown to perform on par with standard deep learning models. Methods extending the Neural-Bag-of-Features model to exploit the temporal domain of the time-series data, as well as long- and short-term information were proposed in \cite{passalis2019deep,passalis2019tetci,Passalis2019a}. Based on the fact that time-series data can be encoded as tensors, the Temporal Attention-Augmented Bilinear (TABL) network was proposed in \cite{Tran2019a}. TABL network was shown to perform on par with state-of-the-art deep learning models in stock mid-price direction prediction, while being much more efficient compared to the rest of the models.

In this paper, we propose a low-rank approximation of the building block of the TABL network, i.e. the TABL layer, to further reduce its computational complexity. We show that, when tested on the problem of stock mid-price direction prediction, the proposed Low-Rank Temporal Attention-Augmented Bilinear (LR-TABL) network is able to achieve the same performance levels as the TABL network, while it requires a smaller number of trainable parameters, as well as a lower number of floating point operations during inference. Such characteristics can make the proposed LR-TABL network a solution that better suits in ultra high-frequency financial time-series analysis problems. 

The remainder of paper is organized as follows. The related works in financial time series analysis are briefly presented in Section \ref{sec:Related_works}, followed by the description of the TABL network in Section \ref{sec:TABL}. The proposed LR-TABL network is described in Section \ref{sec:Proposed_method}. Experimental results are provided in Section \ref{sec:Experiments}, and Section \ref{sec:Conclusions} concludes our work. 

\section{Related work}\label{sec:Related_works}
Data-driven methodologies proposed for financial time series analysis in general, and LOB(Limit Order Book)-based stock mid-price prediction in particular, can be divided into two major categories, i.e. those built on statistical models and data driven models. Methodologies based on statistical models study the statistical properties of the instances forming the time-series data \cite{Bouchaud2002}. A comprehensive survey on statistical methods and stochastic modeling for LOB data can be found in \cite{Cont2011}, and a mathematical study of the order book as a multidimensional continuous-time Markov chain is presented in \cite{Abergel2013}. The underlying assumption of statistical models used for time-series data analysis is that the data is generated by a linear process \cite{Cavalcante2016}. Such assumption limits their ability to model real world data, as these are highly nonlinear and non-stationary.

Methodologies based on data driven models are commonly formed by two building blocks, i.e. a time-series data representation block and a regression or classification block. The time-series data representation block is usually formed by two processing steps, i.e. feature extraction and feature learning. In the feature extraction step the raw time-series data is pre-processed and transformed into feature vectors. Handcrafted features encoding the statistics of the time-series at a short-term and at a long-term window are extracted from multi-dimensional time-series in \cite{Kercheval2015}. Handcrafted features based on technical and quantitative analysis and time-insensitive and time-sensitive indicators are proposed in \cite{Ntakaris2019ssrn}, while an extensive list of econometric features is proposed in \cite{Ntakaris2019}. Using such time-series representations, feature learning can be conducted by using subspace learning techniques, like Principal Component Analysis \cite{788121}, Linear Discriminant Analysis \cite{abdi2010principal}, or Auto-Encoder networks \cite{hinton2006reducing}. A data-driven time-series representation scheme jointly optimizing the feature representation step based on soft vector quantization and the feature learning step based on Clustering-based Discriminant Analysis was proposed in \cite{iosifidis2012multidimensional}. Handcrafted time-series representations like those described above are combined with standard classification models. The Support Vector Machine classifier was used in \cite{zhang2001multiresolution,Huang2005,Kercheval2015}, while in \cite{1257413} a Support Vector Machine classifier with adaptive parameters was shown to perform better than the standard one. A Multilayer Perceptron was used in \cite{zhang2001multiresolution} to predict future bonds returns, while in \cite{Tran2019}, a Heterogeneous Multilayer Generalized Operational Perceptron was shown to build high-performing compact neural network topologies.

Methods combining the feature extraction, feature learning and classification steps in an end-to-end learning process have been recently proposed for financial time-series forecasting, and have shown to outperform methodologies in which these steps are independently designed. A deep learning methodology based on Convolutional Neural Networks for prediction of mid-price changes is proposed in \cite{Tsantekidis2017}. A combined model based on Long Short Term Memory (LSTM) and Convolutional Neural Network (CNN) layers outpeforming individual LSTM and CNN models is proposed in \cite{Tsantekidis2018}. A model combining CNN layers for feature extraction and LSTM layers to capture time dependencies is proposed in \cite{Zhang2019}, and it was shown to outperform most of existing models for mid-price prediction. The Temporal Logistic Neural Bag-of-Features model was proposed in \cite{Passalis2019} and it was shown to perform better than the baseline models. A trainable normalization scheme which can be jointly optimized with any deep learning method for time-series classification was proposed in \cite{Passalis2019a}. The Temporal Attention-Augmented Bilinear (TABL) network was proposed in \cite{Tran2019a} and combines bilinear projections with an attention mechanism. It was shown in \cite{Tran2019a} that TABL network provides state-of-the-art performance while requiring a low number of parameters compared to the competing methods.

\section{Temporal Attention-Augmented Bilinear Network}\label{sec:TABL}
\begin{figure}[htb]
\includegraphics[width=0.95\linewidth]{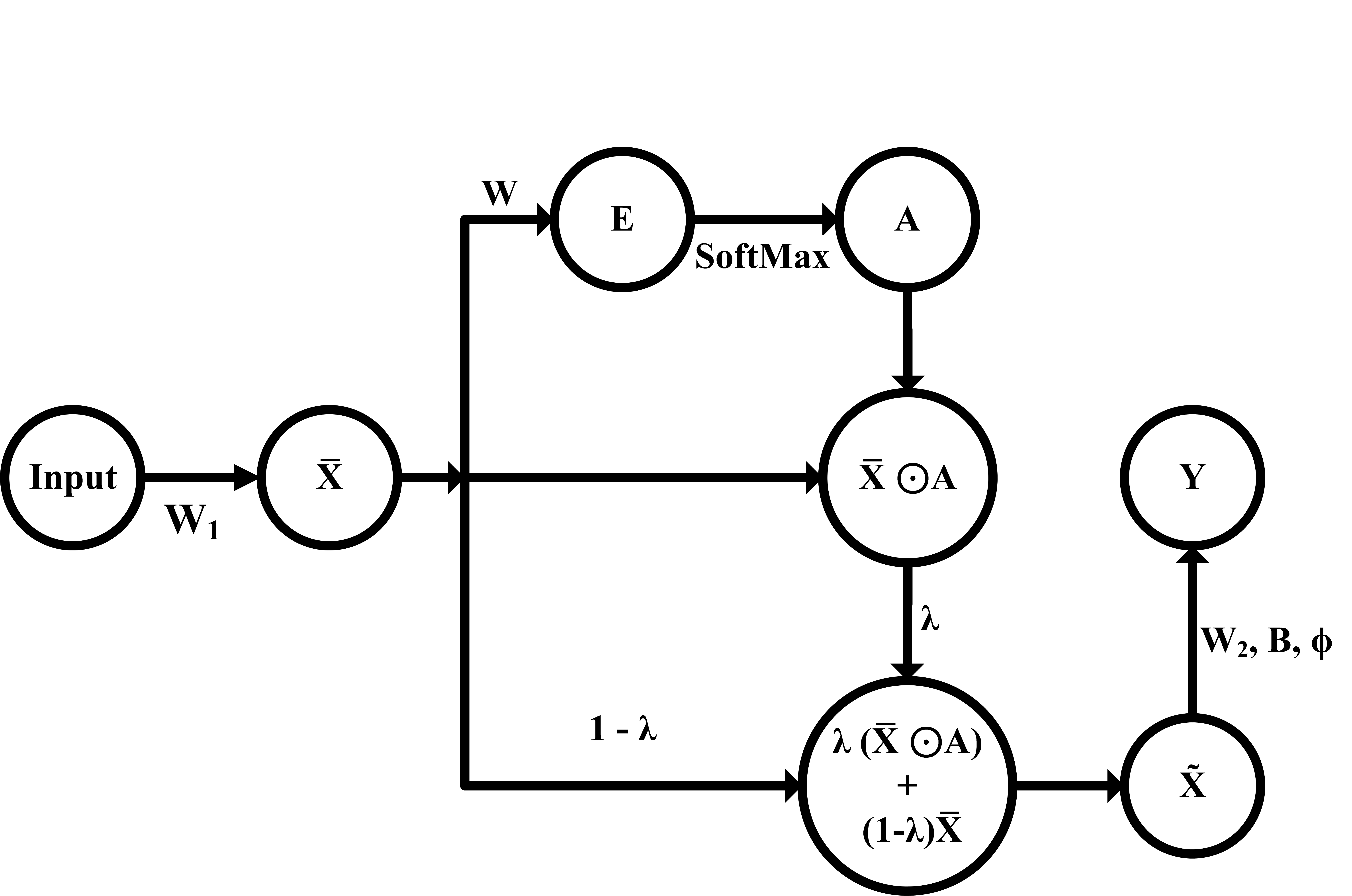}
\caption{The TABL layer based on \cite{Tran2019a}}
\label{fig:TABL_layer}
\end{figure}

The TABL network \cite{Tran2019a} treats the time-series as a second order tensor $\mathbf{X} \in \mathbb{R}^{D \times T}$, where $T$ is the number of time-instances in the time-series. The building block of the network is a bilinear mapping based on parameters $\mathbf{W}_1 \in \mathbb{R}^{D' \times D}$ which are used for applying data transformation, $\mathbf{W}_2 \in \mathbb{R}^{T \times T'}$ which are used for aggregating information at the time domain, and $\mathbf{B} \in \mathbb{R}^{D' \times T'}$ which are used as a bias term. Moreover, another parameter matrix $\mathbf{W} \in \mathbb{R}^{T \times T}$ is used to incorporate an attention mechanism in the bilinear projection. The processing steps applied in the TABL layer are illustrated in Figure \ref{fig:TABL_layer}, and are the following:
\\[2ex]
\noindent{\bf 1: Input data transformation}\\
A new time-series representation $\bar{\mathbf{X}} \in \mathbb{R}^{D' \times T}$ is obtained by
\begin{equation}
\bar{\mathbf{X}} = \mathbf{W}_1 \mathbf{X}.
\end{equation}

\noindent{\bf 2: Attention calculation}\\
The matrix $\mathbf{W}$ (the diagonal elements of which are fixed to a value of $1/T$) is used to encode the relative importance of element each element of $\bar{\mathbf{X}}$, i.e. the importance of $\bar{\mathbf{x}}_{ij}$ with respect to other elements $\bar{\mathbf{x}}_{ik}, \: k \neq j$. This is done by calculating:
\begin{equation}
    \mathbf{E} = \bar{\mathbf{X}} \mathbf{W}
\end{equation}
and introducing the elements of $\mathbf{E}$ in a softmax function, i.e.:
\begin{equation}
    \alpha_{ij} = \frac{ exp(e_{ij}) }{ \sum_{k=1}^T exp(e_{ik}) }.
\end{equation}
The values $\alpha_{ij}$ are used to form the attention mask $\mathbf{A}$.
\\[2ex]
\noindent{\bf 3: Application of the attention mechanism}\\
A soft attention mechanism is applied to the new time-series representation $\bar{\mathbf{X}}$ using a learnable parameter $0 \le \lambda \le 1$ is applied as follows:
\begin{equation}
    \tilde{\mathbf{X}}= \lambda (\bar{\mathbf{X}} \odot \mathbf{A}) + (1-\lambda) \bar{\mathbf{X}}.
\end{equation}

\noindent{\bf 4: Aggregation of temporal information}\\
The final step is to aggregate information in the temporal domain and introducing the result in a (nonlinear) activation function:
\begin{equation}
    \mathbf{Y} = \phi\left( \tilde{\mathbf{X}} \mathbf{W}_2 + \mathbf{B} \right).
\end{equation}

\section{Proposed Low-Rank Bilinear Layer}\label{sec:Proposed_method}
The proposed LR-TABL network is defined as a low-rank bilinear layer with soft attention. Restricting the parameters of the layer to be of low rank lowers the number of the model's parameters, leading to a more efficient online operation. By defining the model as a low-rank model which is trained in an end-to-end fashion (instead of training the full model and applying the low-rank approximation as a post-processing step) we allow its parameters to adapt and achieve performance at the same levels with the full model.

\begin{figure}[htb]
\includegraphics[width=0.9\linewidth]{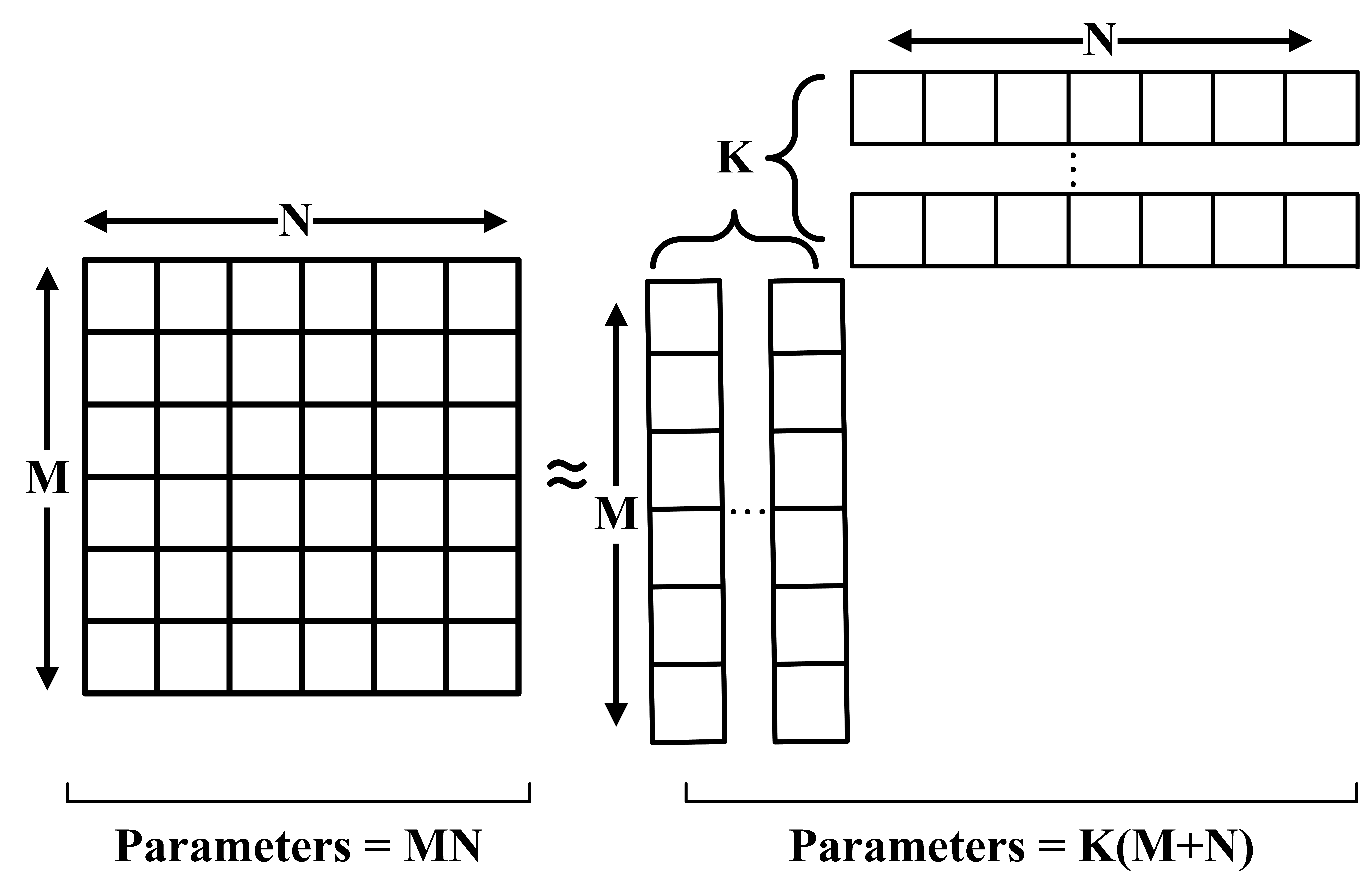}
\caption{Example of low-rank approximation \cite{Sidiropoulos2017}}
\label{fig:low_rank}
\end{figure}

The low-rank approximation of a (2nd-order) tensor $\mathbf{Q} \in \mathbb{R}^{M \times N}$ can be obtained by the multiplication of two tensors $\mathbf{H} \in \mathbb{R}^{M \times K}$ and $\mathbf{V} \in \mathbb{R}^{K \times N}$ \cite{Sidiropoulos2017}:
\begin{equation}
    \mathbf{Q} \approx \mathbf{H} \times \mathbf{V}. \label{Eq:LRapproximation}
\end{equation}
When the rank of $\mathbf{Q}$ is equal to $K$, the exact decomposition of $\mathbf{Q}$ can be calculated by its singular value decomposition and \eqref{Eq:LRapproximation} is changed to an equality. Figure ~\ref{fig:low_rank} illustrates the use of tensors $\mathbf{H}$ and $\mathbf{V}$ to approximate $\mathbf{Q}$ and the corresponding latent space.

The LR-TABL layer models the parameters of the bilinear layer as follows:
\begin{align}
    \mathbf{W}_1 &\rightarrow \mathbf{L}_1 \mathbf{R}_1 \\
    \mathbf{W}_2 &\rightarrow \mathbf{L}_2 \mathbf{R}_2 \\
    \mathbf{W} &\rightarrow \mathbf{L} \mathbf{R}
\end{align}
where $\mathbf{L}_1 \in \mathbb{R}^{D' \times K}$, $\mathbf{R}_1 \in \mathbb{R}^{K \times D}$, $\mathbf{L}_2 \in \mathbb{R}^{T \times K}$, $\mathbf{R}_2 \in \mathbb{R}^{K \times T'}$, $\mathbf{L} \in \mathbb{R}^{T \times K}$ and $\mathbf{R} \in \mathbb{R}^{K \times T}$. Thus, the processing steps of the LR-TABL layer are transformed to the following ones:
\begin{align}
    \bar{\mathbf{X}} = \mathbf{L}_1 \left( \mathbf{R}_1 \mathbf{X} \right) \\
    \mathbf{E}       = \left(\bar{\mathbf{X}} \mathbf{L}\right) \mathbf{R} \\
    \alpha_{ij}      = \frac{ exp(e_{ij}) }{ \sum_{k=1}^T exp(e_{ik}) } \\
    \tilde{\mathbf{X}} = \lambda (\bar{\mathbf{X}} \odot \mathbf{A}) + (1-\lambda) \bar{\mathbf{X}}\\
    \mathbf{Y}       = \phi\left( \left(\tilde{\mathbf{X}} \mathbf{L}_2\right) \mathbf{R}_2 + \mathbf{B} \right)
\end{align}
Here we should note that by using a value of $\lambda = 0$ the proposed layer corresponds to a low-rank bilinear layer (LR-BL). We will combine LR-BL and LR-TABL layers to form the LR-TABL network structures, as described in Section \ref{sec:Experiments}.
\\
[1ex]
\noindent{\bf Time and Space Complexity}\\
Tables \ref{table:Parameters} and \ref{table:TimeComplexity} provide the number of parameters of the LR-TABL layer and time complexity of the processing steps of the proposed LR-TABL layer defined using low-rank tensor approximation. As can be seen, by defining the parameter matrices using a low rank, i.e. when $K \ll D$ and $K \ll T$, the number of parameters and the time complexity of the LR-TABL layer are much lower compared to a TABL layer with the same dimensions.  

\begin{table}[!t]
\caption{Layer number of parameters}\label{table:Parameters}
\begin{center}
\begin{tabular}{|l|c|c|} \cline{2-3}
\multicolumn{1}{c|}{} &   TABL  &   LR-TABL   \\ \hline
$\mathbf{W}_1$        &  $DD'$  &  $(D+D')K$  \\ \hline
$\mathbf{W}_2$        &  $TT'$  &  $(T+T')K$  \\ \hline
$\mathbf{W}$          &  $T^2$  &    $2TK$    \\ \hline
$\mathbf{B}$          &  $TT'$  &    $TT'$    \\ \hline
\end{tabular}
\end{center}
\end{table}
{\renewcommand{\arraystretch}{1.2}%
\begin{table}[!t]
\caption{Layer time complexity}\label{table:TimeComplexity}
\begin{center}
\begin{tabular}{|l|c|c|} \cline{2-3}
\multicolumn{1}{c|}{} &   TABL    &   LR-TABL     \\ \hline
$\bar{\mathbf{X}}$    &  $D'DT$   & $(D+D')KT$    \\ \hline
$\mathbf{E}$          &  $D'T^2$  &  $2D'KT$      \\ \hline
$\mathbf{Y}$          &  $D'T'T$  &  $(T+T')KD' + D'T'$  \\ \hline
\end{tabular}
\end{center}
\end{table}}

\begin{figure*}
\includegraphics[height=4.4cm]{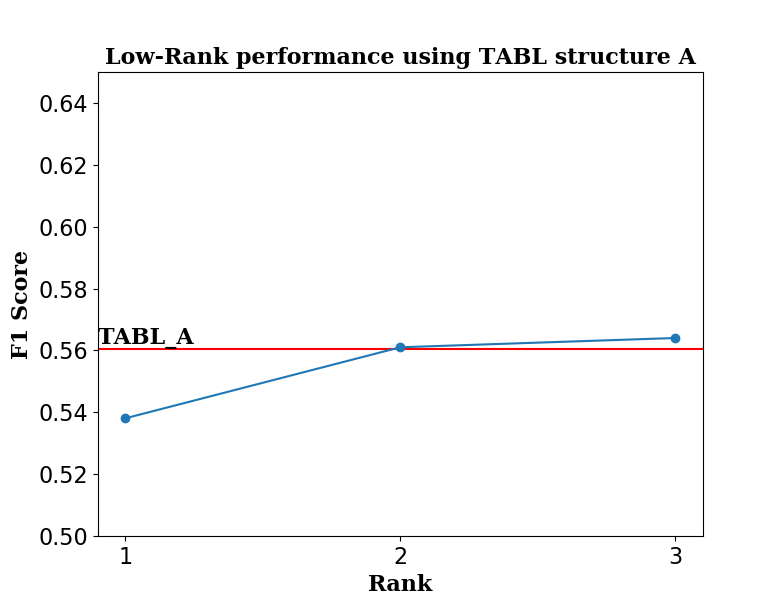}
\includegraphics[height=4.4cm]{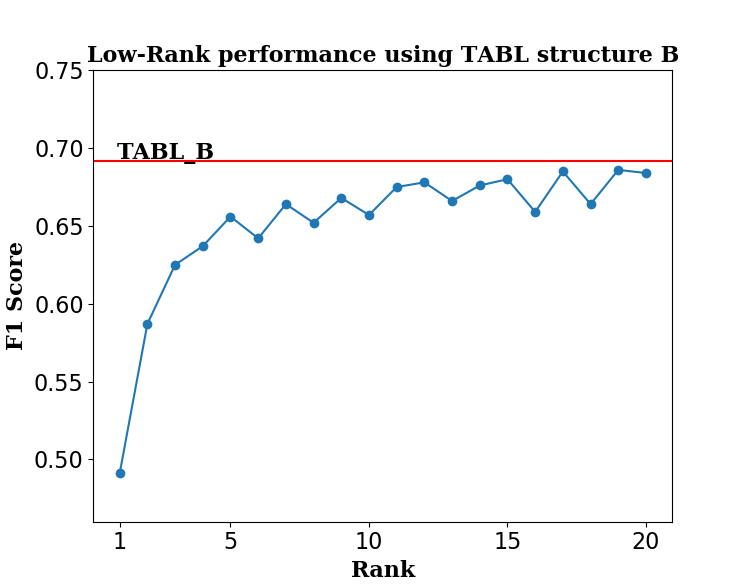}
\includegraphics[height=4.4cm]{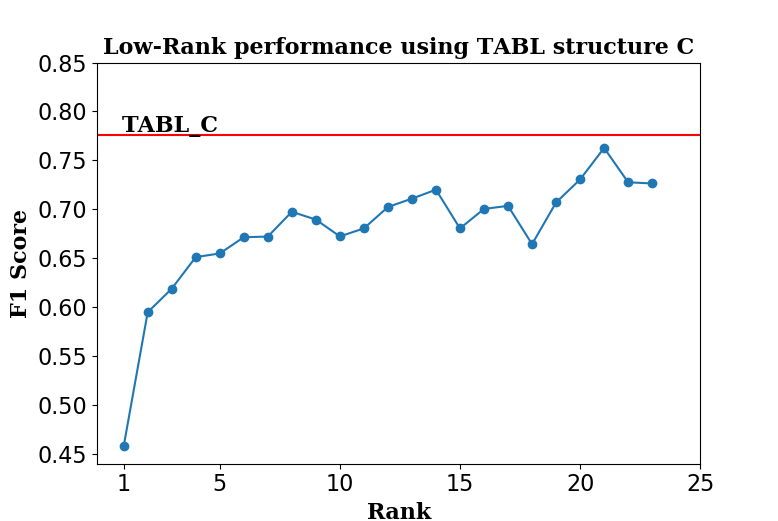}
\caption{F1 Score for TABL A (left), TABL B (middle) and TABL C (right) structures.}
\label{fig:F1_score_TABL}
\end{figure*}
    
\section{Experiments}\label{sec:Experiments}
To evaluate the performance of the proposed approach, we conducted experiments on the publicly available FI-2010 data set \cite{Ntakaris2017}. It consists of ultra high-frequency Limit Order Book data of $5$ Finnish stocks in NASDAQ Nordic for $10$ consecutive days. The database defines an experimental protocol in which time-series data of the first $7$ days are used for training, while the data of the remaining $3$ days are used for evaluation of the performance of the methods. We provide the evaluation of our proposed method using the z-score normalized data.

We follow the same experimental settings as in \cite{Tran2019a}, and compare the performance of a bilinear network based on the proposed LR-TABL layer to that of a bilinear network based on the TABL layer. That is, we use as input the raw Limit Order Book data having ten levels of limit orders and we use the ten most recent instances to form the time-series, i.e. input to the networks is $\mathbf{X} \in \mathbb{R}^{40 \times 10}$. We introduce this time-series to the two networks and we predict the direction in the mid-price ('up', 'stationary', 'down') at prediction horizon of $H = 10$. To have better understanding of performance of the proposed approach, we use the same network structures used in \cite{Tran2019a}, which are the following:
\\[1ex]
\noindent{\bf LR-TABL A:} we have one LR-TABL layer with output size $(3\times 1)$. This layer is followed by a softmax function.
\\[1ex]
\noindent{\bf LR-TABL B:} we use 2 layers, the fist is a LR-BL layer having an output size of $(120\times 5)$ followed by a ReLU activation function. This layer is followed by a LR-TABL layer with output size of $(3\times 1)$ followed by a softmax activation function.
\\[1ex]
\noindent{\bf LR-TABL C:} we use 3 layers, the first two being LR-BL layers with output sizes of $(60\times 10)$ and $(120\times 5)$, respectively, and each followed by a ReLU activation function. The last layer is a LR-TABL layer with output size of $(3\times 1)$ followed by a softmax activation function. 

Since both TABL and the LR-TABL networks predict a class probability vector and the classes forming the classification problem are imballanced, the weighted entropy loss is used:
\begin{equation}
    L = - \sum_{c=1}^C \frac{\epsilon}{N_c} t_c log(y_c),
\end{equation}
where $C = 3$ is the number of classes, $N_c$, $t_c$, $y_c$ are the number of samples, the true probability and the predicted probability of the c-th class, and $\epsilon = 1e6$ is a constant used to ensure numerical stability.  

\begin{table}[!t]
\setlength\tabcolsep{5pt}
\caption{Performance of network structure C}
\label{table:TABL_C_Result}
\begin{center}
\begin{tabular}{|c|c|c|c|c|c|}
\hline
\textbf{$K$}                & \textbf{Acc}            & \textbf{P}              & \textbf{R}              & \textbf{F1}             & \textbf{\# Params} \\ \hline
\textit{\textbf{TABL C}} & \textit{\textbf{0.847}} & \textit{\textbf{0.769}} & \textit{\textbf{0.784}} & \textit{\textbf{0.776}} & 11344              \\ \hline
\multicolumn{6}{|c|}{\textbf{LR-TABL C}} \\ \hline
1           & 0.504  & 0.472  & 0.552  & 0.458 & 1658 \\ \hline
\textit{2}  & 0.716  & 0.584  & 0.609  & 0.595 & 2106 \\ \hline
3           & 0.715  & 0.597  & 0.659  & 0.619 & 2554 \\ \hline
4           & 0.756  & 0.634  & 0.676  & 0.651 & 2879 \\ \hline
5           & 0.758  & 0.638  & 0.678  & 0.655 & 3204 \\ \hline
6           & 0.769  & 0.653  & 0.697  & 0.671 & 3504 \\ \hline
7           & 0.773  & 0.657  & 0.691  & 0.672 & 3804 \\ \hline
8           & 0.791  & 0.681  & 0.718  & 0.697 & 4104 \\ \hline
9           & 0.779  & 0.669  & 0.718  & 0.689 & 4404 \\ \hline
10          & 0.767  & 0.654  & 0.702  & 0.672 & 4704 \\ \hline
11          & 0.779  & 0.665  & 0.699  & 0.680 & 4984 \\ \hline
12          & 0.789  & 0.682  & 0.729  & 0.702 & 5264 \\ \hline
13          & 0.794  & 0.691  & 0.737  & 0.711 & 5544 \\ \hline
14          & 0.804  & 0.703  & 0.742  & 0.720 & 5824 \\ \hline
15          & 0.778  & 0.665  & 0.701  & 0.680 & 6104 \\ \hline
16          & 0.790  & 0.682  & 0.724  & 0.700 & 6384 \\ \hline
17          & 0.794  & 0.687  & 0.724  & 0.703 & 6664 \\ \hline
18          & 0.766  & 0.650  & 0.685  & 0.664 & 6944 \\ \hline
19          & 0.796  & 0.691  & 0.727  & 0.707 & 7224 \\ \hline
20          & 0.810  & 0.712  & 0.754  & 0.731 & 7504 \\ \hline
\textbf{21}                  & \textbf{0.834}          & \textbf{0.748}          & \textbf{0.780}          & \textbf{0.763}          & \textbf{7784}      \\ \hline
22          & 0.807  & 0.710  & 0.750  & 0.728 & 8064 \\ \hline
23          & 0.807  & 0.709  & 0.748  & 0.726 & 8344 \\ \hline
\end{tabular}
\end{center}
\end{table}

\begin{table}[!t]
\setlength\tabcolsep{1pt}
\caption{Performance of network structure A}
\label{table:TABL_A_Result}
\begin{center}
\begin{tabular}{|c|c|c|c|c|c|c|c|c|c|c|}
\hline
\multicolumn{6}{|c|}{\textbf{LR-TABL A}} &
  \multicolumn{5}{c|}{\textbf{TABL A}} \\ \hline
\textbf{$K$} &
  \textbf{Acc} &
  \textbf{P} &
  \textbf{R} &
  \textbf{F1} &
  \textbf{\#Param} &
  \textbf{Acc} &
  \textbf{P} &
  \textbf{R} &
  \textbf{F1}&
  \textbf{\#Param}\\ \hline
1 &
  0.673 &
  0.532 &
  0.542 &
  0.537 &
  78 &
  \multirow{3}{*}{\rotatebox[origin=c]{90}{0.7013}} &
  \multirow{3}{*}{\rotatebox[origin=c]{90}{0.5628}} &
  \multirow{3}{*}{\rotatebox[origin=c]{90}{0.5826}} &
  \multirow{3}{*}{\rotatebox[origin=c]{90}{0.5603}} &
  \multirow{3}{*}{\rotatebox[origin=c]{90}{234}}\\ \cline{1-6}
  2 & 0.681 & 0.548 & 0.577 & 0.559 & 141 &
   &
   &
   &
   &
   \\ \cline{1-6}
   \textit{\textbf{3}} &
  \textit{\textbf{0.695}} &
  \textit{\textbf{0.556}} &
  \textit{\textbf{0.571}} &
  \textit{\textbf{0.563}} &
  \textit{\textbf{204}} &
  &  &  &  & \\ \hline
\end{tabular}
\end{center}
\end{table}

\begin{table}[!t]
\setlength\tabcolsep{5pt}
\caption{Performance of network structure B}
\label{table:TABL_B_Result}
\begin{center}
\begin{tabular}{|c|c|c|c|c|c|}
\hline
\textbf{$K$}                & \textbf{Acc}            & \textbf{P}             & \textbf{R}              & \textbf{F1}             & \textbf{\# Params} \\ \hline
\textit{\textbf{TABL B}} & \textit{\textbf{0.789}} & \textit{\textbf{0.68}} & \textit{\textbf{0.712}} & \textit{\textbf{0.692}} & 5844               \\ \hline
\multicolumn{6}{|c|}{\textbf{LR-TABL B}} \\ \hline
1           & 0.562  & 0.484  & 0.555  & 0.490 & 918  \\ \hline
\textit{2}  & 0.695  & 0.569  & 0.616  & 0.586 & 1226 \\ \hline
3           & 0.734  & 0.608  & 0.648  & 0.624 & 1534 \\ \hline
4           & 0.744  & 0.621  & 0.658  & 0.636 & 1719 \\ \hline
5           & 0.760  & 0.638  & 0.678  & 0.655 & 1904 \\ \hline
6           & 0.754  & 0.631  & 0.653  & 0.641 & 2064 \\ \hline
7           & 0.767  & 0.648  & 0.684  & 0.663 & 2224 \\ \hline
8           & 0.750  & 0.630  & 0.682  & 0.651 & 2384 \\ \hline
9           & 0.766  & 0.648  & 0.692  & 0.667 & 2544 \\ \hline
10          & 0.752  & 0.635  & 0.688  & 0.656 & 2704 \\ \hline
11          & 0.775  & 0.659  & 0.693  & 0.674 & 2864 \\ \hline
12          & 0.777  & 0.661  & 0.696  & 0.677 & 3024 \\ \hline
13          & 0.763  & 0.647  & 0.693  & 0.665 & 3184 \\ \hline
14          & 0.775  & 0.660  & 0.693  & 0.675 & 3344 \\ \hline
15          & 0.778  & 0.664  & 0.698  & 0.679 & 3504 \\ \hline
16          & 0.765  & 0.647  & 0.670  & 0.658 & 3664 \\ \hline
17          & 0.786  & 0.673  & 0.697  & 0.684 & 3824 \\ \hline
18          & 0.772  & 0.658  & 0.668  & 0.663 & 3984 \\ \hline
\textbf{19}                  & \textbf{0.783}          & \textbf{0.670}         & \textbf{0.704}          & \textbf{0.685}          & \textbf{4144}      \\ \hline
20          & 0.783  & 0.671  & 0.697  & 0.683 & 4304 \\ \hline
\end{tabular}
\end{center}
\end{table}

We evaluate our proposed approach based two factors; fist being the performance, and second being the model's efficiency based on number of trainable parameters, compared to the corresponding baseline network. Tables ~\ref{table:TABL_A_Result}-\ref{table:TABL_C_Result} provide the performance obtained by using different values of $K$ based on accuracy, precision, recall, macro F1-score and number of trainable parameters in the model. Here we should note that since the classification problem is formed by imbalanced classes, we consider the F1-score as the metric used in our comparisons. To have a better understating of the relation between the performance achieved by using the different values of $K$ to obtain the low-rank tensor approximation of the network's parameters, Figure \ref{fig:F1_score_TABL} illustrates the F1-score as a function of the value of $K$. As can be observed, the bilinear networks based on the proposed LR-TABL layer performs on par with the the networks using TABL layers, while requiring a lower number of parameters for all network structures.

\section{Conclusion}\label{sec:Conclusions}
This paper proposed a low-rank tensor approximation of the TABL network. We evaluated the proposed LR-TABL network on the the problem of stock mid-price direction prediction problem of the FI-2010 benchmark dataset and compared its performance and efficiency with those of the original TABL network. Results show that we can get the same levels of performance compared to the original TABL network, while reducing the number of trainable parameters, and thus increasing efficiency. The obtained results are an indication that low-rank tensor approximation can be used in other state-of-art deep learning models proposed for financial time-series forecasting to increase their efficient.

\section*{Acknowledgement}
The research received funding from the Independent Research Fund Denmark project DISPA (Project Number: 9041-00004).

\end{document}